**Deep Learning for Classification of Inflammatory Bowel Disease Activity in Whole Slide Images of Colonic Histopathology**

Amit Das[1], Tanmay Shukla, MS[2], Naofumi Tomita, MS[2], Ryland Richards, MD[3], Laura Vidis, MD[3], Bing Ren, MD, PhD[3], Saeed Hassanpour, PhD[1,2,4]

[1]Department of Computer Science, Dartmouth College, Hanover, NH 03755, USA
[2]Department of Biomedical Data Science, Geisel School of Medicine at Dartmouth, Hanover, NH 03755, USA
[3]Department of Pathology and Laboratory Medicine, Dartmouth-Hitchcock Medical Center, Lebanon, NH 03756, USA
[4]Department of Epidemiology, Geisel School of Medicine at Dartmouth, Hanover, NH 03755, USA

[*] **Corresponding Author**: Saeed Hassanpour, PhD
Postal address: One Medical Center Drive, HB 7261, Lebanon, NH 03756, USA
Phone: (603) 646-5715
Email: Saeed.Hassanpour@dartmouth.edu

**Running Title:** Deep learning for grading IBD activity

**Conflicts of Interest**

The authors have no financial, professional, or personal conflicts of interest.

**Funding Sources**

This research was supported in part by grants from the US National Library of Medicine, R01LM012837 (S.H.) and R01LM013833 (S.H.), and the US National Cancer Institute, R01CA249758 (S.H.).

Text Pages: 12; Tables: 2; Figures: 5

# ABSTRACT

Grading inflammatory bowel disease (IBD) activity using standardized histopathological scoring systems remains challenging due to resource constraints and inter-observer variability. In this study, we developed a deep learning model to classify activity grades in hematoxylin and eosin-stained whole slide images (WSIs) from patients with IBD, offering a robust approach for general pathologists. We utilized 2,077 WSIs from 636 patients treated at Dartmouth-Hitchcock Medical Center in 2018 and 2019, scanned at 40× magnification (0.25 μm/pixel). Board-certified gastrointestinal pathologists categorized the WSIs into four activity classes: inactive, mildly active, moderately active, and severely active. A transformer-based model was developed and validated using five-fold cross-validation to classify IBD activity. Using HoVerNet, we examined neutrophil distribution across activity grades. Attention maps from our model highlighted areas contributing to its prediction. The model classified IBD activity with weighted averages of 0.871 [95% Confidence Interval (CI): 0.860-0.883] for the area under the curve, 0.695 [95% CI: 0.674-0.715] for precision, 0.697 [95% CI: 0.678-0.716] for recall, and 0.695 [95% CI: 0.674-0.714] for F1-score. Neutrophil distribution was significantly different across activity classes. Qualitative evaluation of attention maps by a gastrointestinal pathologist suggested their potential for improved interpretability. Our model demonstrates robust diagnostic performance and could enhance consistency and efficiency in IBD activity assessment.

**INTRODUCTION**

Ulcerative colitis (UC) and Crohn's disease (CD) are common chronic inflammatory bowel diseases affecting the colon. Globally, there are approximately 7 million cases of idiopathic inflammatory bowel disease (IBD), with more than 2 million of these cases occurring in the United States alone. The prevalence rates are among the highest in Western countries, such as the US and the UK. In the US, the prevalence rate is 721 cases per 100,000 population, or nearly 1 in 100 Americans. The incidence of new cases continues to rise, with an estimated 39,000 to 56,000 new diagnoses annually in the US alone[1]. This growing prevalence has resulted in a significant economic burden. The annual cost of care for IBD in the US is estimated to be between $14.6 and $31.6 billion, including direct medical costs and indirect costs such as lost productivity [2].

The examination of histopathological slides by pathologists is a critical step in diagnosing and treating many diseases, including IBD. However, the process is time-consuming and resource-intensive with significant interobserver variability among pathologists [3]. Therefore, the development of an automated tool to assist pathologists in assessing the histopathology slides will not only be cost-effective but also improve objectivity and reproducibility.

The widespread availability of digitized whole slide images (WSIs) has significantly advanced computational pathology. Deep learning (DL) methods, such as convolutional neural networks (CNN), have enabled automated diagnosis, prognostication of various diseases and cancers, identification of novel biomarkers, and prediction of clinical endpoints such as disease remission and response to therapy [4,5,6,7]. However, the large size of WSIs, often exceeding billions of pixels, challenges the computational efficiency of conventional DL models such as CNNs on graphic processing units (GPUs). To address the limitation, techniques such as multiple instance learning and attention mechanism have been used to improve performance [8].

Recently, transformers, which utilize self-attention mechanism, have emerged as a significant development in DL to effectively capture contextual information for input data. Inspired by their success in large language models such as GPT, vision transformers (ViTs) are increasingly used for computer vision tasks in various domains, including WSI analysis. They have consistently shown improved performance compared to earlier methods [9,10,11]. Among them, MaskHIT, a methodology for WSI classification and survival prediction using masked pre-training for histology images with a transformer model, showed improved accuracy for both cancer type classification and survival prediction while maintaining a low computational burden due to stochastic sampling [11].

Deep learning (DL) methods are increasingly used in IBD for diagnosis, assessment of endoscopic and histological activity, and predicting clinical outcomes with models analyzing WSIs of colonic mucosal biopsies to study pathological patterns, inflammatory cell distribution, and histological activity [12,13]. Models have used CNNs to predict histological activity regarding the Nancy scoring index in patients with UC [14], ViTs to predict tissue appearance and endoscopic scores [15], and DL to analyze eosinophilic density and clinical outcomes [16]. Other models reference validated scoring systems such as Geboes Score, Nancy Histological Index, Robarts Histopathology Index, and PICaSSO for predicting histological activity [17,18,19]. Recently, CNN-based models have shown high accuracy in predicting histological activity and remission [19]. However, the use of ViT-based models for classifying histopathological activity in IBD remains limited.

Despite these advancements, most DL models for predicting histological activity in IBD patients focus on validated histological scores, often using binary classification to distinguish between the presence and absence of histological remission. This emphasis is primarily because controlled clinical trials in IBD require correlation with standardized histological activity scores. However, grading histopathological activity in IBD remains challenging outside these trials due to time and resource constraints, lack of trained pathologists, and significant inter-observer variability. Consequently, despite the development of over 30 histological scoring indices, their adoption in routine practice is minimal, and community pathologists typically use a semi-objective scale of activity [20,21].

Given these challenges, recent research has focused on the correlation of neutrophilic infiltration with inflammatory activity in IBD. A simplified scoring index based on neutrophil presence has proven efficient in predicting histological remission and may be considered as the minimum standard for assessment of inflammation. This neutrophil-only score reduces subjectivity, requires no additional training for pathologists, and is suitable for developing CNN-based DL models for automated grading [22,23]. However, the neutrophil-only assessment of inflammation in IBD is relatively new and has not been widely validated. Professional societies and expert panels recommend use of validated scoring systems such as Robarts and Nancy index, which have been shown to correlate well with important clinical outcomes such as relapse, hospitalization, need for therapeutic escalation, use of corticosteroid therapy, and long-term development of dysplasia and colon cancer [24,25,26]. Considering that the adoption of these histological scoring indices in clinical practice has remained minimal, in this study, our primary objective is to develop and evaluate a DL model for classifying the activity of IBD using high-resolution H&E-stained WSIs into four levels: inactive, mildly active, moderately active, and severely active based on colitis activity. These categories are intentionally designed to be intuitive and simpler than the more complex and detailed scoring systems like the Robarts and Nancy indices, which, while validated and correlated with clinical outcomes, require extensive knowledge and training and can be time-consuming to apply in routine practice [20,21]. Our work represents the first effort to develop a DL model that classifies grades of histopathological activity in WSIs of tissue slides from IBD patients, suitable for use by general pathologists who may not depend on formal histological scoring systems. Additionally, we developed a pipeline to quantify neutrophil infiltration, revealing statistically meaningful differences across various grades of colitis activity. This analysis offers an alternative perspective on the dataset, complementing the classification of inflammatory activity and enhancing our understanding of the underlying pathology. Finally, to further support pathologists, we provide visualizations of the attention maps generated by our transformer model to enhance interpretability and assist in the clinical decision-making process.

## MATERIALS AND METHODS

**Dataset**

The dataset in this study is from colonoscopic biopsy slides with H&E–stained formalin-fixed, paraffin-embedded (FFPE) samples obtained from IBD patients who were treated at the Dartmouth-Hitchcock Medical Center, a tertiary academic medical center in Lebanon, NH, in the years 2018 and 2019. We collected all available slides for each patient and all the slides were digitized at a magnification of 40× (0.25 μm/pixel) using a Leica Aperio AT2 scanner (Leica Biosystems, Wetzlar, Germany). This dataset comprises 2,077 whole slide images (WSIs) from 636 patients. These WSIs were reviewed and categorized into four classes based on colitis chronicity and activity, as determined by board-certified pathologists with subspecialty training in gastrointestinal pathology involved in our study. The assessment of colitis chronicity and activity is based on the presence of several histological features, such as chronic inflammatory infiltrate in lamina propria (including lymphocytes and eosinophils), basal plasmacytosis, mucin depletion, crypt architecture alteration, Paneth cell metaplasia; neutrophils in the lamina propria, surface and crypt epithelium (cryptitis), and within the lumen of crypts (crypt abscesses); presence of erosions and ulcers[25]. The use of this dataset in our study was approved by the Dartmouth Health Institutional Review Board. Detailed information about the dataset is provided in Table 1.

| Class | Number of WSIs (% of total) | Number of Patients (% of total) | Number of Slides per Patient (mean ± std) |
|---|---|---|---|
| Inactive | 885 (43%) | 191 (33%) | 3.4 ± 2.4 |
| Mildly active | 647 (31%) | 350 (55%) | 2.5 ± 2.2 |
| Moderately active | 358 (17%) | 187 (29%) | 1.9 ± 1.4 |
| Severely active | 187 (9%) | 115 (18%) | 1.6 ± 0.9 |
| Total | 2,077 (100%) | 636 (100%) | 3.3 ± 2.6 |

**Table 1:** A summary of the dataset, categorizing 2,077 WSIs from 636 patients into four IBD classes: inactive, mildly active, moderately active, and severely active. Note that patients may have slides from separate colonic segments belonging to different classes.

**Preprocessing**

Our preprocessing step utilized an open-sourced library for digital whole slides to remove background and extract foreground (i.e., tissue) regions from WSIs [27]. From the foreground regions within each WSI, patches of size 224 x 224 at a magnification of 20x were extracted. This magnification was selected to balance capturing detailed histological morphology with the computational cost of analyzing high-resolution images. The final patches for our analysis were selected based on a threshold of containing at least 5% tissue, ensuring that even patches from the edges of a tissue included significant content. These patches were then passed through a ResNet34 model to obtain their embeddings, which were used in our transformer framework. We recorded the coordinates of each patch as metadata, which were used in the subsequent stages of positional encoding in our transformer, as well as for visualization.

**Development of IBD Classifier for WSIs**

Figure 1 illustrates the pipeline used for classifying the WSIs into four categories: inactive, mild, moderate, and severe. We leveraged the MaskHIT pipeline, a state-of-the-art transformer-based framework for analyzing WSIs in computational pathology, in this work [11]. The MaskHIT architecture is based on a ViT with 8 heads and 12 layers, chosen for its capability to capture complex patterns in high-dimensional data. Average pooling was used as the fusion method to integrate features from different regions of the WSIs. A weighted loss function was implemented to effectively manage the class imbalance, preventing model bias toward the more frequent classes.

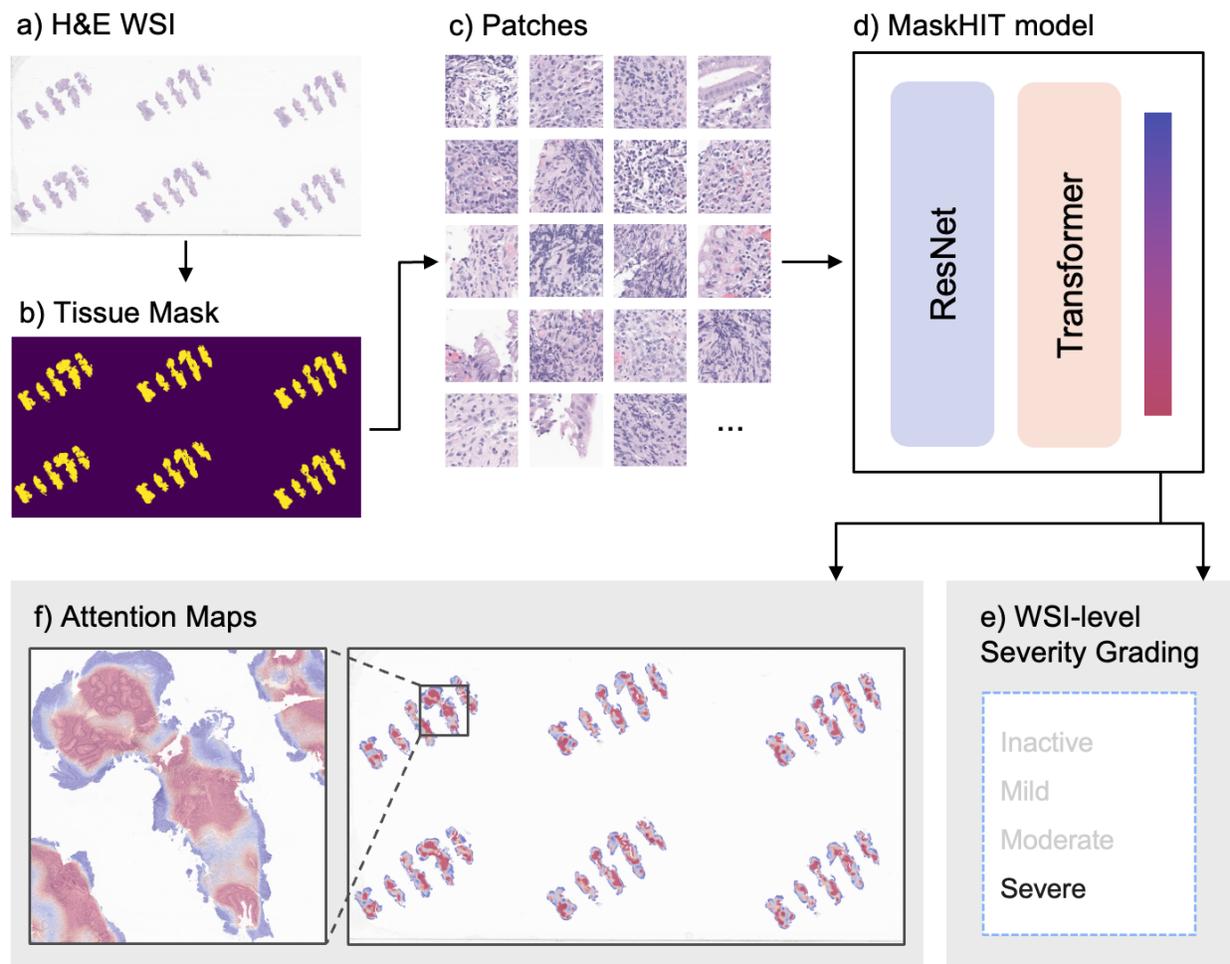

**Figure 1:** The figure illustrates the pipeline used for classifying histopathological images into four categories: inactive, mild, moderate, and severe. The process begins with an H&E whole slide image (WSI) (a), from which a tissue mask is generated (b). The WSI is then divided into smaller patches (c) that are processed through a ResNet34-based patch feature extractor. These features are subsequently fed into the MaskHIT transformer model (d), which outputs an aggregated CLS token representing the slide-level feature vector used for the final classification into the predefined activity categories (e). Attention maps (f) are generated to visualize the model's focus and help interpret the grading during the classification process.

We used a five-fold cross-validation strategy in our model development while ensuring that each fold was evenly distributed and that all slides from the same patient were kept within the same fold. During validation, up to 175 regions from each WSI were sampled to ensure robust performance assessment across varied tissue regions. At each training iteration, the model was trained using only four regions from a single WSI to prevent overfitting. Training included early stopping with a patience threshold of 10 epochs, guided by the weighted F1 score on the validation set. We used the Adam optimizer, default parameters for weight decay and learning rates from MaskHIT pipeline, a batch size of 16, and a dropout rate of 0.1.

In this work, we used a pre-trained model of MaskHIT, which was trained on patches of size 448×448 pixels at a magnification of 20x on a subset of The Cancer Genome Atlas (TCGA) database. The fine-tuning process on the IBD dataset was carried out on an NVIDIA RTX 6000 Ada GPU, with a total training time of approximately 8 hours. Memory consumption peaked at 20 GB. Deep learning packages used in this pipeline includes PyTorch (version 2.1.0) [28] and torchvision (version 0.16.0) [29].

**Cellular Composition Analysis**

To investigate the cellular composition of IBD tissue samples of different grades of activity, we used the HoVerNet pipeline [30] to examine the distribution of cell types, specifically neutrophils. HoVerNet is a deep learning model that simultaneously segments and classifies nuclei within tissue histology images. It uses horizontal and vertical distances of nuclei to perform segmentation, followed by a separate branch to classify the nuclei. We used the HoVerNet model with the MoNuSac checkpoint for segmentation and classification, categorizing cells into four types: epithelial cells, lymphocytes, macrophages, and neutrophils [31].

To perform the HoVerNet analysis, WSIs were processed at their original 40x magnification level. These high-resolution images were analyzed using the HoVerNet model, and visualizations of regions within the WSIs were created to show the boundaries of different cell types. An expert pathologist collaborator in our study reviewed these visualizations for quality control and confirmed the accuracy of the HoVerNet for identifying cell types on our slides. Following this qualitative validation, the counts of neutrophils, lymphocytes, and macrophages were quantified to investigate their distribution across various activity levels of colitis.

**Statistical Analysis**

For the MaskHIT-based model, performance was assessed using aggregated weighted AUC (Area Under the receiver operating characteristic Curve), precision, recall, F1 score, and confusion matrices across testing sets. Additionally, 95% confidence intervals (95% CI) were calculated using bootstrapping based on the model's predictions and included in brackets for all the metrics to ensure the reliability of the performance metrics. For the HoVerNet analysis, the Mann-Whitney U-test was performed to generate U-statistics, effect sizes, p-values, and probabilities indicating whether the distributions of cell counts were significantly different across the varying activity levels. These probabilities were calculated using the Probability of Superiority (PS) measure, which represents the likelihood that a randomly selected cell count from a higher activity level is greater than a randomly selected cell count from a lower activity level.

**RESULTS**

**ViT Model Performance**

The Vision Transformer achieved an aggregated weighted AUC of 0.871 [0.860-0.883], precision of 0.695 [0.674-0.715], recall of 0.697 [0.678-0.716], and F1 score of 0.695 [0.674-0.714] in predicting inflammatory activity of IBD (Table 2). The confusion matrix is shown in Figure 2, and the AUC is depicted in Figure 3.

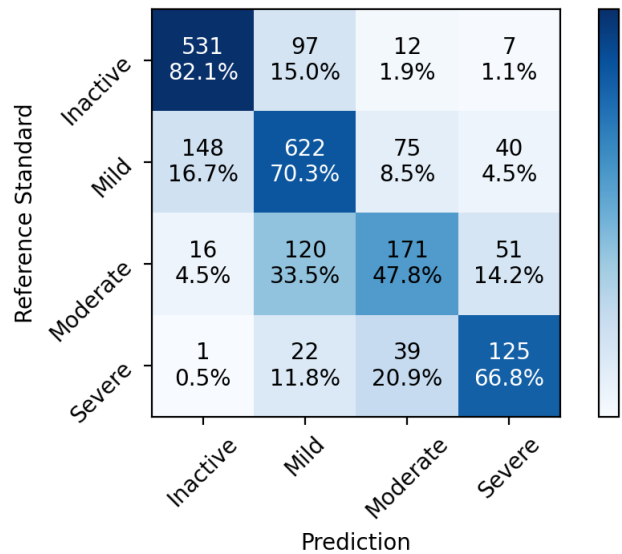

**Figure 2:** The confusion matrix shows the model's performance by depicting the counts and percentages of true labels against predicted labels across the four classes over the testing set. Each cell contains the number of WSIs classified into each category and the percentage relative to the true labels' total. The matrix highlights the model's performance in correctly classifying each class and the extent of misclassifications. Inactive, mild, and severe classes show strong true positive rates, while moderate cases have a lower rate, indicating an area for potential model improvement.

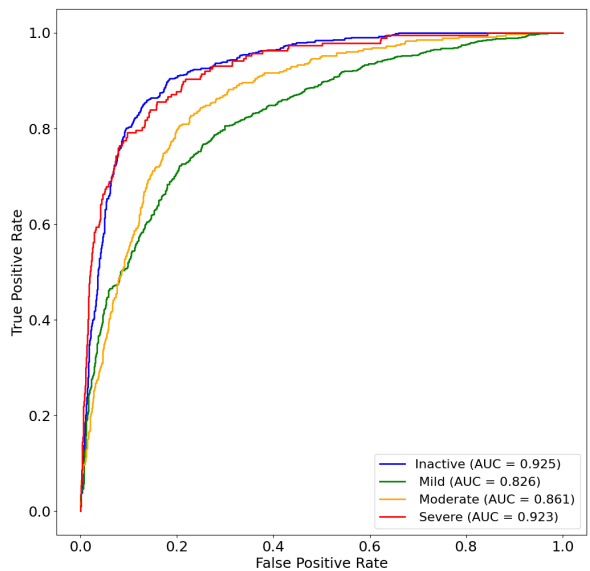

**Figure 3:** The ROC curves illustrate the model's performance in classifying WSIs into four categories: inactive, mild, moderate, and severe. Each curve represents the true positive rate (sensitivity) plotted against the false positive rate (1-specificity) for different threshold values. The aggregated Area Under the Curve (AUC) is provided for each class across the testing set.

The model demonstrated strong performance in classifying inactive and mildly active cases. Specifically, for inactive cases, 531 out of 647 were correctly classified, yielding an accuracy of 82.1% [95% CI, 0.789-0.850]. For mildly active cases, 622 out of 885 were correctly classified, resulting in an accuracy of 70.3% [95% CI, 0.672-0.733]. In classifying moderately active cases, the model achieved moderate performance, with 171 out of 358 correctly classified, resulting in an accuracy of 47.8% [95% CI, 0.425-0.531]. However, 120 moderately active cases were incorrectly misclassified as mildly active, indicating some level of overlap between these classes. The model showed good performance in classifying severely active cases, with 125 out of 187 correctly classified, achieving an accuracy of 66.8% [95% CI, 0.596-0.735].

| Class | AUC | Precision | Recall | F1 |
|---|---|---|---|---|
| Inactive | 0.925 | 0.763 | 0.821 | 0.791 |
| Mildly active | 0.826 | 0.722 | 0.703 | 0.712 |
| Moderately active | 0.861 | 0.576 | 0.478 | 0.522 |
| Severely active | 0.923 | 0.561 | 0.668 | 0.609 |
| Weighted Average | 0.871 [0.860-0.883] | 0.695 [0.674-0.715] | 0.697 [0.678-0.716] | 0.695 [0.674-0.714] |

**Table 2:** This table presents the performance metrics of our ViT model across four categories: inactive, mildly, moderately, and severely active. The metrics include AUC, Precision, Recall, and F1 scores per class, and average (weighted) values.

**Distribution of neutrophils**

Figure 4 illustrates the distribution of neutrophil cell counts using violin plots, showing a significant association with disease activity. A Mann-Whitney one-sided U-test revealed statistically significant differences: mild vs. inactive ($U = 331,001.5, p = 1.47 \cdot 10^{-7}, r = 0.153$), moderate vs. mild ($U = 221,408.5, p = 3.33 \cdot 10^{-28}, r = 0.396$), and severe vs. moderate ($U = 47,504.5, p = 4.52 \cdot 10^{-16}, r = 0.419$).

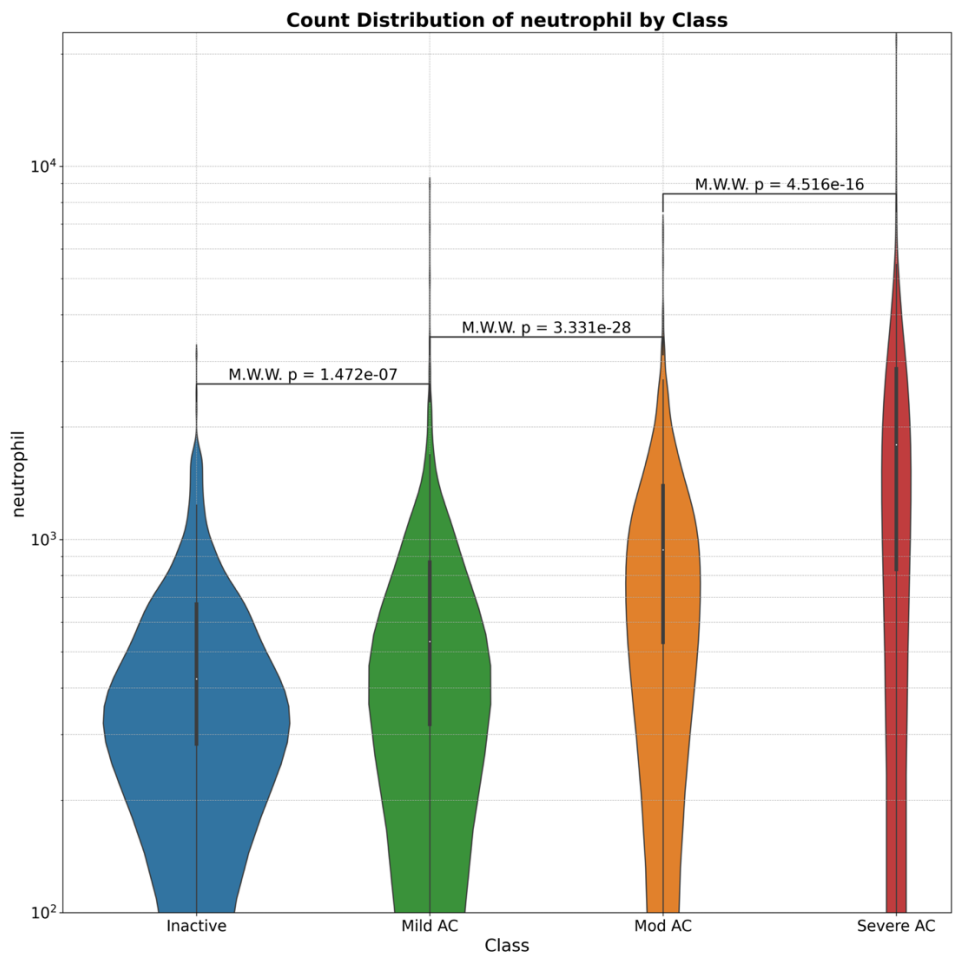

**Figure 2:** Violin plots representing the distributions of neutrophils segmented and classified using HoVerNet inference with the MoNuSAC checkpoint across the full dataset of WSIs. Plots are shown for each class: inactive, mild, moderate, and severe. The distributions exhibited notable differences, with statistical significance (Mann-Whitney-Wilcoxson test) observed between the various classes.

To further quantify these differences, we calculated the probabilities that a randomly selected neutrophil count from a group with higher activity is greater than one from a lower activity group: mildly active vs. inactive (57.7%), moderately vs. mildly active (69.8%), and severely vs. moderately active (71.0%). These values highlight the increasing trend in neutrophil counts with disease activity, supporting the statistical significance observed in the Mann-Whitney U-tests.

**Visualization**

Attention maps of the fine-tuned model were generated to analyze model predictions across the inactive, mildly, moderately, and severely active categories. The process involved examining the attention scores region by region. For each region, the attention weights of the classification (CLS) token were extracted with respect to every patch within that region. These attention weights were then used to create heatmaps, visualizing the attention distribution. The heatmaps were overlaid on the corresponding regions of the original images to provide a visual representation of the model's focus.

Figure 5 presents randomly selected examples of attention overlays, including the original patches and patches overlaid with HoVerNet annotations, providing a comprehensive view of the model's focus areas and the impact of fine-tuning.

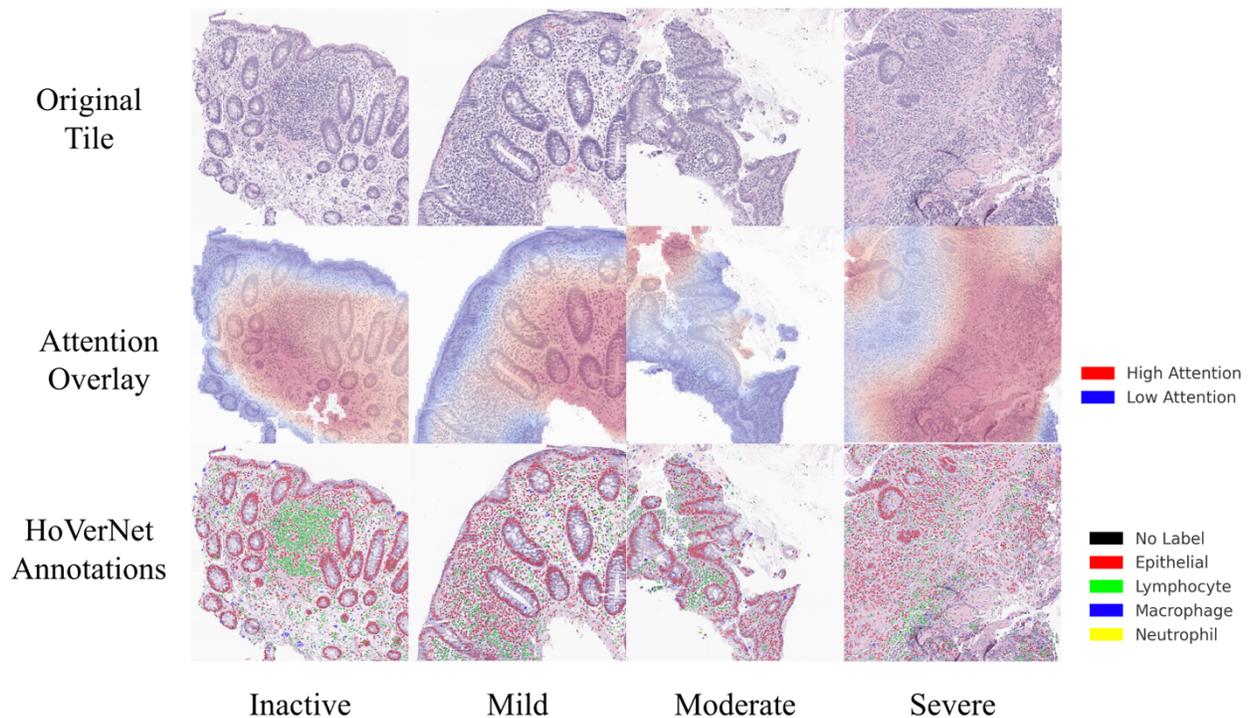

**Figure 3:** Comparison of tissue regions, attention maps, and HoVerNet annotations across different activity levels of colitis shown from left to right as inactive, mild, moderate, and severe. The top panel comprises representative images of tissue regions for each class. The second panel displays attention map overlays showing the fine-tuned attention weights for the corresponding regions. Red corresponds to higher attention values, blue to lower values. The bottom panel shows HoVerNet annotations overlaid on the corresponding tissue images. Cell types are color-coded as follows: red (epithelial cells), blue (macrophages), yellow (neutrophils), green (lymphocytes), black (no label).

## DISCUSSION

Previous studies on applying DL technology for grading IBD inflammatory activity have primarily focused on the automated prediction of a variety histopathological scores, such as the Geboes Score, Nancy Histological Index, and Robarts Histopathology Index [17,19]. These scores are mainly used in research and therapeutic drug trials and are adopted minimally in community practice [20]. Building on this foundation, the primary objective of this study is to develop and evaluate DL models for classifying IBD inflammatory activity that can be applied in a clinical setting for day-to-day patient management. Our decision to classify IBD activity into four distinct and clinically relevant categories—inactive, mildly, moderately, and severely active—reflects a more practical approach that aligns with current clinical practices by providing clear, actionable grades that can be easily interpreted by general pathologists.

In our study, the ViT model, trained using the MaskHIT pipeline, demonstrated robust performance across the four activity levels. The aggregated weighted F1 score of 0.695 [0.674-0.714] and an AUC of 0.871 [0.860-0.883] highlight the model's ability to distinguish between different activity levels of UC effectively. An automated prediction tool for grading the activity of inflammation in IBD patients could simplify patient management, especially in a resource-constrained environment. Of note, the majority of our model's mistakes, according to the confusion matrix, involve the classification of moderate cases as mild. This reflects some overlap in the morphology of these classes and the clinical difficulty in differentiating between the two.

Given the recent interest in grading activity and predicting clinical outcomes in colitis based on the neutrophil-only assessment of histological inflammation, we also studied the neutrophil counts in WSI with different grades of activity. Integrating HoVerNet to analyze cell counts provided additional insights into the cellular composition across different grades of colitis. Although not the project's primary focus, this analysis serves as a valuable complement to the primary model predictions and sanity check, revealing that neutrophil counts significantly increased with disease activity, peaking in the severely active class, and providing some insight into the underlying mechanism of this classification. Of note, we also experimented with a logistic regression model trained with HoVerNet neutrophil counts, which yielded suboptimal performance and therefore was not included in this manuscript. This experiment highlights that while neutrophil counts offer some degree of separability between inactive and severely active classifications, they are insufficient as a standalone feature for reliable classification for other activity levels. Thus, the ViT model's enhanced performance underscores the importance of utilizing advanced deep learning architectures for this task.

The statistical distribution of activity classes in our dataset reflects the varying degrees of activity in our dataset, providing a diverse and balanced set of examples for the model to learn from, which is a strength of the study. Such diversity is crucial for training robust and effective models capable of accurately classifying colitis activity in new, unseen WSIs. Unlike the currently published AI models that use deep learning (DL) technology for the assessment of digital pathology slides specific to UC[16–19], our model is applicable to both of the common types of IBD, namely UC and CD. Another strength of our model is that, while other published DL models can only distinguish between remission and inflammation[19], our model is capable of classifying grades of inflammatory activity in IBD and enhancing risk stratification, which may directly impact management decisions, including the need for escalated treatment strategies.

There are some limitations to our predictive model. Other published computer-aided diagnosis (CAD) systems for UC have demonstrated good correlation with endoscopic activity and the ability to predict the risk of clinical flare[16,19]. However, in our study, we did not have access to that information. In the future, in addition to cross validating our model with established histopathological and endoscopic scoring systems, we plan to assess its ability to predict clinically important outcomes, such as clinical flares, the need for surgical interventions, and, importantly, the risk of dysplasia development.

The ViT model shows a clear ability to accurately distinguish between inactive and severely active IBD cases, demonstrating its strength in identifying the extreme ends of the disease spectrum. Attention visualization maps further revealed the model's decision-making process, enhancing interpretability. These findings underscore the potential of advanced deep learning models in histopathological image analysis for the automated classification of IBD activity according to widely used categories, which may improve clinical care for patients with IBD. Future work includes exploring further model optimization, incorporating additional data sources to enhance performance, and conducting multi-institutional prospective evaluations to demonstrate the generalizability of our approach.

**ACKNOWLEDGMENTS**

We thank Michael Winter for his advice on the early stages of our research, providing useful clinical insights.